\documentclass[10pt,journal]{IEEEtran}
\usepackage{amsmath,amsfonts}
\usepackage{algorithmic}
\usepackage{algorithm}
\usepackage{array}
\usepackage[caption=false,font=normalsize,labelfont=sf,textfont=sf]{subfig}
\usepackage{textcomp}
\usepackage{stfloats}
\usepackage{url}
\usepackage{verbatim}
\usepackage{graphicx}
\usepackage{cite}
\usepackage{multirow}
\usepackage{color}
\usepackage{xspace}

\makeatletter
\DeclareRobustCommand\onedot{\futurelet\@let@token\@onedot}
\def\@onedot{\ifx\@let@token.\else.\null\fi\xspace}

\def\eg{\emph{e.g}\onedot} 
\def\ie{\emph{i.e}\onedot}

\def\etal{\emph{et al}\onedot}
\makeatother

\begin{document}

\title{{PSNet}: Fast Data Structuring for Hierarchical \\ Deep Learning on Point Cloud}

\author{{Luyang}~{Li},~
        {Ligang}~{He},~\IEEEmembership{Member,~IEEE,}
        {Jinjin}~{Gao}~
        and~{Xie}~{Han}
\thanks{{Luyang}~{Li} is with School of Data Science and Technology, North University of China, Taiyuan 030051, China, and also with Shanxi Information Industry Technology Research Institute Co., Ltd., Taiyuan 030012, China (e-mail: lly007@live.cn).}
\thanks{{Ligang}~{He} is with Department of Computer at the University of Warwick, Coventry, CV4 7AL, United Kingdom  (e-mail: ligang.he@warwick.ac.uk).}
\thanks{{Jinjin}~{Gao} is with experimental center, Shanxi University of Finance and Economics, Taiyuan 030006, China  (e-mail: 20141005@sxufe.edu.cn).}
\thanks{{Xie}~{Han} is with School of Data Science and Technology, North University of China, Taiyuan 030051, China (e-mail: hanxie@nuc.edu.cn).}
\thanks{Copyright © 2022 IEEE. Personal use of this material is permitted. However, permission to use this material for any other purposes must be obtained from the IEEE by sending an email to pubs-permissions@ieee.org.}
}



\maketitle

\begin{abstract}
   In order to retain more feature information of local areas on a point cloud, local grouping and subsampling are the necessary data structuring steps in most hierarchical deep learning models. Due to the disorder nature of the points in a point cloud, the significant time cost may be consumed when grouping and subsampling the points, which consequently results in poor scalability. This paper proposes a fast data structuring method called PSNet (Point Structuring Net). PSNet transforms the spatial features of the points and matches them to the features of local areas in a point cloud. PSNet achieves grouping and sampling at the same time while the existing methods process sampling and grouping in two separate steps (such as using FPS plus kNN). PSNet performs feature transformation pointwise while the existing methods uses the spatial relationship among the points as the reference for grouping. Thanks to these features, PSNet has two important advantages: 1) the grouping and sampling results obtained by PSNet is stable and permutation invariant; and 2) PSNet can be easily parallelized. PSNet can replace the data structuring methods in the mainstream point cloud deep learning models in a plug-and-play manner. We have conducted extensive experiments. The results show that PSNet can improve the training and inference speed significantly while maintaining the model accuracy.
\end{abstract}



\section{Introduction}
\label{sec:intro}

\IEEEPARstart{P}{oint} cloud is a common 3D data format. It has been widely used in areas such as robotics and autonomous driving. Since a point cloud has a non-Euclidean data structure \cite{bronstein2017geometric}, it is still facing significant challenges to apply deep learning methods directly to point clouds and extracting effective information.

Following on the great success of CNN in images \cite{he2016deep,krizhevsky2017imagenet,simonyan2015very,szegedy2015going,cgn-tcsvt,edcrf-tcsvt}, the voxel model \cite{shabat20183dmfv,brock2016generative,maturana2015voxnet,riegler2017octnet,wang2015voting,wang2017o,zhou2018voxelnet,cicek20163d,hollow-3DR-CNN-tcsvt} extends the 2D CNN and is applied directly to the regularized point cloud data. However, the voxel model loses the resolution of the point cloud, and causes the significant increase in computation cost. 

\begin{figure}[t]
\begin{center}
\includegraphics[scale=0.99]{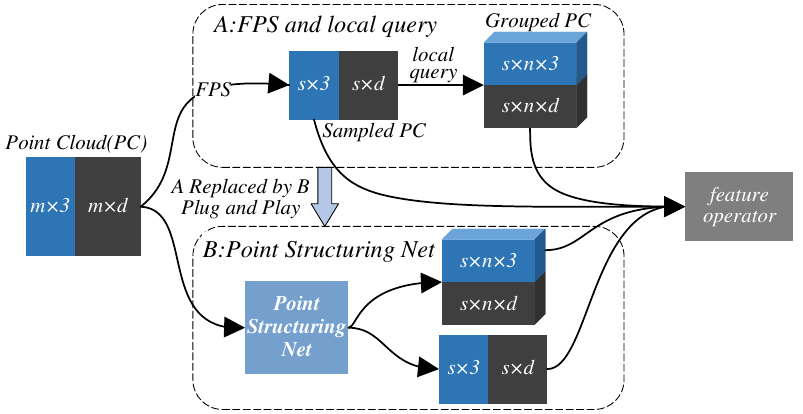}
\end{center}
   \caption{The role of PSNet: replacing FPS and local points query in the hierarchical feature abstraction of local areas. Blue tensors are the sets of the 3D coordinates, black tensors are the sets of abstract features, $m$, $s$, $n$ is the number of points, local areas, and local points respectively. PSNet can replace the traditional data structuring methods easily in the plug-and-play manner.}
\label{fig1}
\end{figure}

The current mainstream methods \cite{qi2017pointnet,charles2017pointnet,transformer3d2021-tcsvt,psr2021-tcsvt,komarichev2019a,li2018pointcnn,liu2019densepoint,liu2019relation,mao2019interpolated,wu2019pointconv,xu2018spidercnn,zhao2019pointweb,chen2019clusternet,shen2018mining,simonovsky2017dynamic,lhsa2021-tcsvt,zhang2018a, msnfe-tcsvt} take the points directly as the input of the network model. Based on the seminal method presented in \cite{charles2017pointnet}, many research studies have been conducted to improve the abstraction of local context features. These works can be divided into three categories: point-based \cite{qi2017pointnet}, spatial convolution \cite{komarichev2019a,li2018pointcnn,liu2019densepoint,liu2019relation,mao2019interpolated,wu2019pointconv,xu2018spidercnn,zhao2019pointweb} and graph convolution \cite{chen2019clusternet,shen2018mining,simonovsky2017dynamic,zhang2018a}. For local context feature abstraction, these methods mainly adopt the hierarchical architecture for local context aggregation. Due to the disorder feature of the points in a point cloud, the points with similar spatial positions are often stored in non-contiguous locations in memory. However, the hierarchical architecture for local context aggregation needs to organize the data of the point cloud according to their spatial feature. This is a typical data structuring problem and has to be handled in all the above three categories of deep learning methods. 

\IEEEpubidadjcol

A current mainstream data structuring method includes two steps as shown in Part A of \figurename~\ref{fig1}: 1) uniform subsampling such as Farthest Point Sampling (FPS) \cite{eldar1997the}, 2) local grouping such as ball query, kNN or cube query. The subsampling step obtains a subset of points in the point cloud, and then the local grouping step uses each sampling point as the center of a local area, and groups the points around the center point into a local area. The grouping methods have to calculate the Euclidean distances between each of the sampled points and all other points to determine which points should be placed in the local area. However, most of the calculations are actually not necessarily needed, and the same calculations are repeated in both steps. Moreover, FPS searches for the sampling points in sequence, which makes it difficult to parallelize the calculations. The study in \cite{liu2019pvcnn} shows that the time spent in data structuring can account for 88\% of the total processing time. 

On the other hand, FPS is unstable in the sense that the sampling result of FPS is closely related to the starting points of the FPS process (which are typically selected randomly) and the point ordering in a point cloud. The unstable sampling results of FPS make the effectiveness of FPS unstable too. In addition, although FPS can ensure the uniformity of sampling, FPS is essentially based on a single metric \ie the Euclidean distance between the points, and does not actively consider other features of the points, which may limit the effectiveness of the sampling results.

Now let us think how human would perform the data structuring work. When human tries to group the points in a point cloud into local areas, the points that are far apart or show unrelated features would be ignored straightaway. Consequently, the calculation of the distances between these points can be avoided. Human is able to sense which points should be grouped together, which are the points with the similar spatial location feature. This insight inspired the Point Structuring Net (PSNet) method proposed in this paper.

PSNet tackles the data structuring issue in point clouds, which accounts for a substantial proportion of the time in both model training and inference due to the non-Euclidean data. PSNet aims to replace the existing data structuring methods in deep networks on point clouds as shown in \figurename~\ref{fig1}. PSNet is novel in the sense that it addresses this issue from a totally different perspective, which significantly reduces the time spent in data structuring. In PSNet, each individual point, represented by its Cartesian coordinates $(x, y, z)$, is transformed to the high-dimensional features by our spatial feature transform function (SFTF), which determines the correlation degree between each point and the abstract features of each local area. Then, the points that are highly correlated with the local abstract features are grouped into a local area. The point with the feature closest to the abstract feature of a local area is regarded as the sampling point of this local area. 

Moreover, we found that the symmetry of the Cartesian coordinates may cause incorrect grouping of the points in the symmetrical parts of a point cloud. Inspired by the spherical coordinates, we add two more parameters as the input spatial feature of a point, which reduces the grouping errors effectively. The data structuring of PSNet is based on the spatial features of the points, not only on a single distance metric. PSNet is embedded into a deep network and co-trained with the network (supervised by the loss function of the original deep network). Consequently, PSNet is adaptable to the objectives of the original learning tasks while encoding the features of the points and the local areas. This is the underlying reason why PSNet is more effective than the heuristic methods such as FPS and kNN.

The existing methods perform subsampling and grouping as two separate stages in sequence (first subsampling and then grouping). They make use of the spatial relations among the points to make subsampling and local grouping decisions. Therefore, they have to calculate the Euclidean distances between the points repetitively, which takes long computing time. In PSNet, the features transformed by SFTF can be used for both sampling and grouping simultaneously, which not only avoids the expensive distance computations, but also reduces unnecessarily repeated computations.

The SFTF can be performed on each point independently. The process can therefore be fully parallelized. PSNet can generate the effective data structuring results with much less time during the model training, while the data structuring time spent by PSNet in model inference can be almost neglected. PSNet can easily replace the data structuring methods in point cloud deep learning models in a plug-and-play manner. We have conducted extensive experiments on a variety of mainstream deep learning models on point clouds. The experimental results show that PSNet is effective and significantly reduces the training and inference time of the models. We have implemented PSNet, which has been open-sourced at \emph{https://github.com/lly007/PointStructuringNet}.

The remainder of this paper is organized as follows. The related work is discussed in section~\ref{sec:RelatedWork}. PSNet is presented in detail in section~\ref{sec:PSNet}. The experimental results are presented in section~\ref{sec:Experiments}. The paper is concluded in section~\ref{sec:Conclusion}.

\section{Related Work}
\label{sec:RelatedWork}

In this section, we discuss various types of data structuring methods in deep learning on point clouds. They mainly include basic but widely used methods, the custom-designed methods for specific models, the random point sampling method and learning-based methods.

\subsection{Most Widely Used Methods: FPS, kNN and Ball Query}

PointNet++ \cite{qi2017pointnet} aggregates the local features hierarchically. Other methods \cite{komarichev2019a,li2018pointcnn,liu2019densepoint,liu2019relation,mao2019interpolated,wu2019pointconv,xu2018spidercnn,zhao2019pointweb} define unique spatial convolution operators, and perform the convolution on the local area.  In addition, there are many graph-based methods \cite{chen2019clusternet,shen2018mining,simonovsky2017dynamic,wang2019graph,zhang2018a}, which use the coordinate-based local-area queries to construct the graphs. All these methods use FPS \cite{eldar1997the} for subsampling and use kNN or ball query for local grouping. However, FPS, kNN and ball query have the issues in both efficiency and effectiveness.

In terms of efficiency, as the number of points in the point cloud increases, the amount of calculations in FPS increases significantly. Moreover, FPS searches for the farthest points iteratively in a point cloud, which is difficult to be parallelized. The mainstream local grouping methods are ball query or kNN. Both methods need to calculate the Euclidean distances between each of the sampled points and all other points. Many distance calculations are repetitively performed in the sampling and the grouping phase. Our studies show that a large portion of these calculations are unnecessary. 

As for the effectiveness, since FPS and kNN (or ball query) are essentially the heuristic approaches (based on the distance between the points), the data structuring results obtained by FPS and kNN (or ball query) are the same for different training models and learning tasks, not adapted to the feature abstraction operators, the network architecture and the objectives of the learning tasks. These static data structuring results limit the effectiveness of the training models.

\subsection{Bespoke Methods for Specific Models}

Some works custom-designed the data structuring methods to suit their unique network architectures. SO-Net \cite{li2018so} subsamples the point clouds via Self-Organizing Map (SOM) and discovers the neighbourhood of the sampled points by kNN. KD-Net \cite{klokov2017escape} organizes the point cloud structure and transforms the features hierarchically through KD-Tree. SPG \cite{spg2018} generates local grouping of the point clouds according to the basic geometric shapes through the Superpoint Graph. But its computational overhead of shape partitioning and supergraph analysis are expensive. PVCNN \cite{liu2019pvcnn} and VoxelNet \cite{zhou2018voxelnet} find the local areas by the voxel. They do not sample the points and therefore cannot reduce the number of input features in the deeper layers of the networks. The architectures of these methods are different from the usual PointNet++-based hierarchical deep networks. Rather, they design the special feature abstraction operators in their training networks in order to aggregate the local features. Therefore, it is hard to plug these data structuring methods into the existing, generic training networks.

\subsection{Random Sampling}

The Random Point Sampling (RPS) finds the sampling points randomly in the point cloud and is therefore very fast. However, the effectiveness of the sampling results is unstable. Consequently, using RPS in a typical method \cite{qi2017pointnet} can significantly reduce the effectiveness of the network. 

Some works \cite{hu2020randla,xu2020grid} use RPS as the subsampling method. However they have to design the special network architecture or feature abstraction operators for the training networks in order to offset the instability caused by RPS. For example, Grid-GCN \cite{xu2020grid} uses the spatial voxel division to constrain RPS and prevent the excessive randomization of sampling. RandLA-Net \cite{hu2020randla} does not have the constraints in the random sampling phase, but has to be mapped to a dilated residual block to counteract the information loss caused by random sampling. These specially designed methods are not universally applicable and cannot be used in most deep learning models for point clouds.

\subsection{Learning-based Methods}

Some works proposed the learning-based sampling methods, which learn the sampling strategies through the lightweight neural networks. SampleNet \cite{lang2020samplenet} and S-NET learn a subset of a point cloud by the neural networks. In these methods, other heuristic sampling methods (such as FPS) have to be used to provide the baseline for learning. Then the sampling points generated by the learning-based methods have to be compared with the points in the original point cloud, which introduces extra computations. CAE \cite{balin2019concrete} and PAT \cite{yang2019modeling} use the reparameterization trick to calculate the sampling weight matrix according to the relationship among all points. CP-Net\cite{nezhadarya2020cpnet} evaluated the contribution to the aggregated features and selected Critical Points (CP) as the sampling points. The learning-based methods are optimized for both the task and the deep network in sampling to improve the effectiveness of the sampling. However, these methods do not perform local grouping. Rather, the traditional local grouping methods such as kNN or ball query have to be used to complete the data structuring. 

PSNet proposed in this work is a fast data structuring method based on deep learning, and adopts a totally different approach from all the data structuring methods discussed above. PSNet does not reply on the spatial relations among the points for data structuring, and can be fully parallelized. PSNet is a generic approach, and can be embedded into the existing mainstream deep learning networks for point clouds in a plug-and-play manner. Moreover, in most methods in the literature, sampling and local grouping are two separate tasks. In PSNet, grouping and sampling are performed at the same time. This feature further improves the efficiency of PSNet.

\section{PSNet}
\label{sec:PSNet}

Querying local areas on a point cloud can be regarded as a multi-clustering problem in the Euclidean distance space (\ie clustering the points into local areas). In a 3D space, the points that are closer to each other should be grouped into a local area. The traditional methods (\eg radius query and kNN) find a class of points close to a certain point from the above perspective and group them together. This type of methods first uses subsampling (\eg FPS) to find a subset of the point cloud as the center points of the neighborhoods. PSNet (Point Structuring Net) proposed in this work does not use the Euclidean distance between the points as the selection criterion for the neighborhood, but exploits a multi-clustering method to divide the points into local areas.

\subsection{Spatial Location Feature Transform Function}

Traditional methods explicitly rely on the heuristic geometric meaning of points such as the Euclidean distance between points. More specifically, they must calculate the distance between a certain point and \textbf{all} remaining points, and then determine their relationship (\ie whether they belong to the same local area). Since a local area is usually small, most of the calculations, which are performed all remaining points, are unnecessary. It is also worth noting that in the grouping methods, whether a point belongs to a local area is only related to its distance to the center of the local area (\ie a sampled point), irrelevant to the local shape. 

The points that are close to each other in the 3D space often have similar features when they are transformed by the same neural network operator. The \emph{Spatial Features Transform Function (SFTF)} proposed in this work does not focus on the heuristic correlation between the points (\eg the Euclidean distance). Rather, \emph{SFTF} abstracts the spatial features of each individual point, and groups the points with similar features in the same local area. 

A point cloud is $P=\left\{\boldsymbol{c_i}\middle| i=1,2,\cdots,m\right\}$, where $m$ is the number of points, $\boldsymbol{c_i}\in \mathbb{R}^d$ is the spatial features of the $i$-th point $p_i$, $L=\left\{l_j\middle| j=1,2,\cdots,s\right\}$ is the set of local areas which $P$ is divided into, $l_j$ is local area $j$, $s$ is the number of local areas and also the number of sampling points.

In \emph{SFTF}, local grouping for a point cloud $P$ is treated as the problem of determining the correlation between a point $p_i$ and one (or more) of the s local areas (\ie whether $p_i$ is a member of a particular local area). SFTF aims to transform the spatial features of a point into the degree of correlation between a point and local areas. 

Define $t(x):\mathbb{R}^d\rightarrow \mathbb{R}^s$ as the \emph{SFTF} function for point $p_i$ with the spatial features $\boldsymbol{c_i}$ (the feature dimension is $d$):

\begin{equation}
\boldmath{f_i}=t\left(\boldmath{c_i}\right)
\label{e1}
\end{equation}
where $\boldsymbol{f_i}\in \mathbb{R}^s$ is the vector of correlation degrees between point $p_i$ and each of $s$ local areas.

With \emph{SFTF}, the problem is converted into an $s$-class classification problem given a correlation vector $\boldsymbol{f_i}$ (\ie given $\boldsymbol{f_i}$ which of the $s$ classes point $p_i$ is classified into).

Eq.~(\ref{e2}) is used to apply the $sigmoid$ function to the correlation vector $\boldsymbol{f_i}$ to obtain the probability vector $\boldsymbol{q_i}$, an element of which holds the probability that point $p_i$ belongs to one of the s local areas $l_j (j=1,2,\cdots,s)$, where $\boldsymbol{q_i}\in Z^s,\ 0 < Z < 1$. For a single classification problem, we only need to obtain the index of the element which has the largest value in the vector (denoted by $argmax{\left(\boldsymbol{q_i}\right)}$). For a multi-class (assuming $n$-class) problem, the elements which have top $n$ values (denoted by ${argtop}_n(\boldsymbol{q_i})$) are the $n$ classes that $p_i$ should be allocated to.

\begin{equation}
\boldsymbol{q_i}=sigmoid(\boldsymbol{f_i})
\label{e2}
\end{equation}

However, after local grouping of a point cloud, the maximum number of points in a local area should be fixed (\ie each area contains at most $k$ points), which cannot be guaranteed by the above formulation of the classification problem. Therefore, we modify the classification process as follows. We first apply Eq.~(\ref{e2}) to each point and then apply Eq.~(\ref{e3}), where $T$ is the pointwise broadcast version of $t$. $T\left(P\right)\in \mathbb{R}^{m\times s}$ takes as input the entire set of points in point cloud $P$ and outputs a two dimensional matrix in which there are $m$ rows (corresponding to $m$ points in $P$) and $s$ columns (each row is $\boldsymbol{f_i}$ in Eq.~(\ref{e1})); $Q\in \mathbb{R}^{m\times s}$ is the membership probability matrix between each point in $P$ and each local area in $L$ (the set of local areas).

\begin{equation}Q=sigmoid\left(T\left(P\right)\right)\label{e3}\end{equation}

A column in $Q$ is denoted by $\boldsymbol{e_j}\in \mathbb{R}^m$. Each element in $\boldsymbol{e_j}$ is the probability that point $p_i$ belongs to local area $l_j$. The indices of the points in area $l_j$ can be obtained through these elements.

$n$ is the number of points in $l_j$. Then finding the indices of the points in $l_j$ can be formulated as:

\begin{equation}{indices}_j={argtop}_n\left(\boldsymbol{e_j}\right)\label{e4}\end{equation}
where ${indices}_j\in \mathbb{R}^n$, $arg{top}_n$ is the indices of the top $n$ elements in the vector.

The above process can be understood in the following way. A local area $l_j$ is abstracted to be a type of feature. The points belonging to $l_j$ should be closer to the abstract feature of $l_j$ after performing the feature transformation function $T$. The process of finding the indices of the top values in $e_j$ is equivalent to finding $n$ points in the point cloud that best match the feature of local area $l_j$.

Our grouping method reflects an important viewpoint of us in performing local grouping for point cloud models. In the existing grouping methods such as ball query and kNN, the grouping decisions are made essentially based on the Euclidean distance, which is a single heuristic metric. We would like to argue that in point cloud models, which is non-Euclidean data structure, the distance should not be the sole metric to determine the grouping of points. Our PSNet can adjust the grouping of points adaptively based on the extracted local features, rather than only on a single heuristic metric such as the Euclidean distance. Actually, even for the Euclidean data structure such as images, some studies have proposed to use adaptive kernels (instead of a kernel with the fixed size such as a 3x3 matrix) to perform the convolution operation. For example, the work in \cite{dai2017deformable, DCNN-TCSVT} proposed the ``Deformable Convolutions", in which the convolution kernel may be deformed adaptively in the learning process and as the result the local divisions of the image are also adjusted adaptively. 

\subsection{Selection of Subsampling Points}

Vector $\boldsymbol{e_j}$ introduced in Eq.~(\ref{e4}) represents the probabilities that point $p_i$ is a member of each local area $l_j (j=1,2,\cdots,s)$. The point with the highest probability is the point whose features best match the features of local area $l_j$ among all points in the local area and therefore should be selected as the subsampling point. Therefore, the subsampling point for local area $l_j$ can be determined by:

\begin{equation}
{sub}_j=\left\{p_k\middle| k=argmax\left(\boldsymbol{e_j}\right)={argtop}_1\left(\boldsymbol{e_j}\right)\right\}
\label{e5}
\end{equation}

It is easy to understand that the output of $arg{sort}_1$ in Eq.~(\ref{e5}) exists in the output of $arg{sort}_n$ in Eq.~(\ref{e4}). Namely, after local grouping is completed, the set of subsampled points can also be determined. $sub\in \mathbb{R}^{s\times3}$ denotes the set of subsampled point cloud.

It is worth noting that in the PointNet paper\cite{qi2017pointnet}, when the authors discuss the reason for the effectiveness of their proposed method, they stated that the points corresponding to the maximum values in the feature channels ``summarize the skeleton of the shape". In the experiment section, we will visualize our sampling results (\figurename~\ref{fig5}), which show that the points sampled by our method form the skeleton of an object.

\subsection{Micro-geometric Meaning}

The micro-geometric meaning of PSNet is that after being transformed by the trained \emph{SFTF}, the points with similar locations are also similar in each channel (a channel represents an abstract feature of the local area) of the $s$-dimensional feature space. Since \emph{SFTF} transforms each point independently and ${argtop}_n$ is a symmetric function, PSNet is permutation invariant. This means that the data structuring results obtained by PSNet are stable. Namely, the results of PSNet are not affected by the point ordering in a point cloud, and PSNet will obtain similar data structuring results for the 3D objects with similar spatial features. In contrast, the results obtained traditional sampling and grouping methods such as FPS and kNN heavily rely on the starting points of the sampling/grouping method. However, the starting points are often randomly selected, which makes the effectiveness of their results unstable. Also due to this problem, the traditional methods may generate different sampling and grouping results for the objects with similar features. 

\subsection{Correct Symmetry Errors}

Although the Cartesian coordinate can fully describe the position of a point in a point cloud, our studies show that if the shape of a point cloud is symmetric, \emph{SFTF} may transform the symmetric points in a point cloud to the similar high dimensional features. This is due to the symmetrical parts of an object often have similar spatial features. For example, when the two symmetric points $[1,2,-3]$ and $[1,2,3]$ will be transformed to the same value after right multiplying by $[1,1,0]^{T} $. In order to address this issue, we introduce two more features, polar angle ($\theta$) and azimuthal angle ($\varphi$) (inspired by the spherical coordinate) as the input spatial feature of a point (in addition to its Cartesian coordinate). The values of $\theta$ and $\varphi$ are supposed to be in the ranges of (0, $\pi$] and (0, $2\pi$]. Therefore, the problem of transforming the symmetric points can be almost eradicated, which is supported by our experimental results. Typically, the data format of a point cloud only provides the Cartesian coordinates of the points. $\theta$ and $\varphi$ required by \emph{SFTF} can be calculated through the Cartesian coordinates, as in Eq.~(\ref{e6}), where $x$,$y$,$z$ are the Cartesian coordinates of a point, $r=\sqrt[]{x^2+y^2+z^2} $ is the radius of the spherical coordinate.

\begin{equation}
\left\{\begin{matrix} 
  \theta=\arccos{\left({z}/{r}\right)} \\  
  \varphi=\arctan{\left({y}/{x}\right)} 
\end{matrix}\right. 
\label{e6}
\end{equation}

However, when $\arctan{\left({y}/{x}\right)}$ in Eq.~(\ref{e6}) is used to calculate $\varphi$,  it has to determine the quadrant of the $xy$ plane. In the implementation, we avoided this computation overhead by invoking the $atan2(y,x)$ function instead (which does not have to determine the quadrant). $atan2(y,x)$ is provided by most mathematical libraries\cite{abadi2016tensorflow,harris2020array,paszke2019pytorch}. Moreover, since the value range of $atan2$ is $(-\pi, \pi]$, we define  $\varphi=\ atan2(y,x)+\pi$, so that the value range of $\varphi$ can become $(0, 2\pi]$. In summary, the final input spatial features of a point are:

\begin{equation}
\begin{split}
    c_i & =\left[x,y,z,\theta,\varphi\right] \\ & =\left[x,y,z,\arccos{\left(z/r\right)},atan2\left(y,x\right)+\pi\right]
\end{split}
\label{e7}
\end{equation}

\subsection{Differentiable Indexing}

The $argmax$ function in Eq.~(\ref{e5}) is non-differentiable, because the indices are a series of discrete integer values. So backpropagation cannot propagate through $argmax$ in training to calculate the gradients of the parameters in the feature transform function. $argmax$ needs to be replaced by a differentiable function \cite{jang2016categorical}.

We apply the $Gumbel\ Softmax$ distribution function to the probability distribution $\boldsymbol{e_j}\in \mathbb{R}^m$ of the sampling points:

\begin{equation}
\boldsymbol{\widetilde{e_j}}=softmax\left(\left(\ln\left(\boldsymbol{e_j}\right)+{noise}_j\right)/temperature\right)
\label{e8}
\end{equation}

where $temperature$ is annealing temperature. ${noise}_j$ can be expressed by Eq.~(\ref{e9}), where $U$ is a uniform distribution.

\begin{equation}
{noise}_j=-\ln\left(-\left(U_i\right)\right),U_i\sim U\left(0,1\right)
\label{e9}
\end{equation}
where $U$ is uniform distribution.

When the annealing temperature value approaches $0$, the $Gumbel\ Softmax$ function tends to convert $\boldsymbol{e_j}$ to a one hot vector. The index of the element with the value of 1 in $\boldsymbol{\widetilde{e_j}}$ is $argmax\left(\boldsymbol{e_j}\right)$. We apply the $Gumbel\ Softmax$ function to all $\boldsymbol{\widetilde{e_j}}$ in matrix $Q$:

\begin{equation}
\widetilde{Q}={GumbelSoftmax}_m (Q)
\label{e10}
\end{equation}
where ${GumbelSoftmax}_m$ is Gumbel Softmax calculated by the column. $\widetilde{Q}\in \mathbb{R}^{m\times s}$ is the sparse matrix of sampling indices. The sampled points can be obtained by left-multiplying the point cloud $P\in \mathbb{R}^{m\times3}$ by $\widetilde{Q}\in \mathbb{R}^{m\times s}$, \ie Eq.~(\ref{e11}), where $sub \in \mathbb{R}^{s\times3}$.

\begin{equation}
sub={\widetilde{Q}}^T\times P
\label{e11}
\end{equation}

\begin{figure}[t]
    \centering
    \includegraphics[scale=1.1]{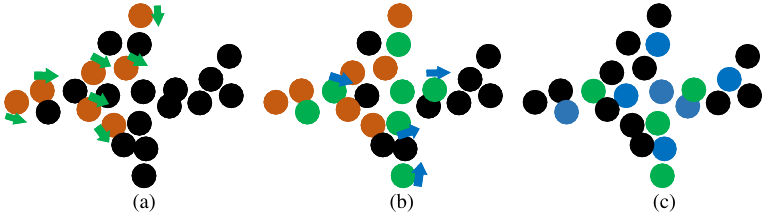}
    \caption{Changes of sampling results during training. Brown, green and blue points respectively represent the sampling results of different training stages. The arrows indicate the deviation between the current sampling points and the target sampling points. Deviations provide directions from current sampling points to target. Green arrows are the deviation from brown points to green points, blue arrows are the deviation from green points to blue points. Blue points are the final sampling result.}
    \label{fig2}
\end{figure}

Differentiable indexing is an important reason for the effectiveness of PSNet. Traditional heuristic data structuring methods rely on uniform sampling, FPS guarantees the uniform distribution of local areas. This idea is very intuitive and vanilla, which has certain universal applicability. But we would like to argue that it is not the best way for data structuring. In fact, recent studies have shown that there are differences in the degree of attention of different regions of the sample data (\eg, attention mechanism \cite{attentionisallyouneed}), and a heuristic method will restrict this difference. In this sense, the adaptive point cloud data structuring method is more effective. We chose \emph{Gumbel} as the index strategy instead of \emph{Reinforce}, because \emph{Gumbel} can provide the guidance for adaptive point cloud sampling to converge in the best way. \figurename~\ref{fig2} shows how \emph{Gumbel} affects the sampling results during the training process. Due to the random initialization of the network, the positions of sampling points in PSNet may be uneven at first, as shown by brown points in the \figurename~\ref{fig2}(a), which is obviously not the best sampling result. However, \emph{Gumbel} will affect the training of the parameters of PSNet, causing a slight deviation of the target sampling point (shown by the arrows in the figure). This deviation will cause a slight change in the sampling target, which is more conducive to task optimization results. The deviation degree depends on the temperature setting. After a period of training, the sampling points will reach the positions of the green points in (\figurename~\ref{fig2}(b)). As the training continues further, the sampling points will reach the positions of the blue points (\figurename~\ref{fig2}(c)). PSNet will be trained at the same time as the task, which means that in addition to optimizing the parameters of the original operator to improve the effectiveness of feature abstraction, the training \textbf{also adjusts the sampling method} to improve the effectiveness of the model from another perspective. 

\subsection{The Architecture of PSNet}

\begin{figure*}[ht]
\begin{center}
\includegraphics[scale=1]{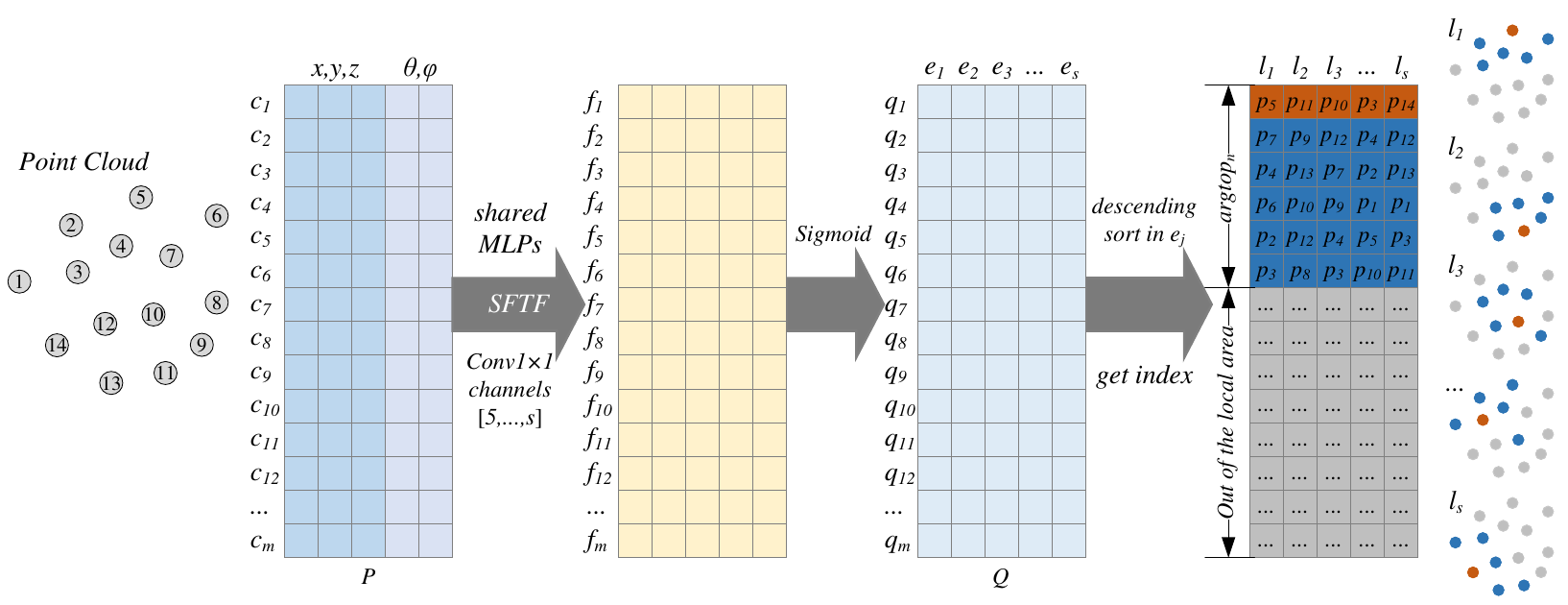}
\end{center}
   \caption{The architecture of PSNet. The spatial location feature transform function is implemented by $1\times1$ convolution, then applies sigmoid to the matrix, sort each column in descending order. In the output matrix of PSNet, the orange points are the subsampling points, and the blue points are the points in the local areas of orange points.}
\label{fig3}
\end{figure*}

PSNet is a sub-network that can be plugged in a deep learning network for point clouds. Since PSNet does not use the position relationship among the points, $T$ can be applied to all points in parallel. This is a main reason why PSNet can speed up the subsampling process significantly over FPS. 

The architecture of the network is shown in the \figurename~\ref{fig3}. The MLP (Multilayer Perception) network is shared among different points to implement the function $T$ (the first step in \figurename~\ref{fig3}). In order to take full advantage of the parallel computing capability of modern GPUs, we implement the MLP by multi-layer $1\times1$ convolution. The number of channels the first layer in the convolution is $5$ as shown in Eq.~(\ref{e7}), while the number of channels of the last layer is the number of sampling points $s$ (which is also the number of local areas). There can be one or more hidden layers in MLP (the impact of the number of layers will be evaluated in Table \ref{t8}). MLP outputs an $m\times s$ matrix. As described in Eq.~(\ref{e2}) and Eq.~(\ref{e4}), $sigmoid$ is applied to the matrix (Step 2: $sigmoid$ in \figurename~\ref{fig3}) and the elements in each column are sorted in the descending order of their values (Step 3 in \figurename~\ref{fig3}). By taking the index of the first element and the indices of the top $n$ elements in each column, we can obtain the sampling point and the group of points in the local area.

Different indexing strategies are adopted in the training and the inference stage to improve the inference speed. In order to ensure the differentiability of the MLP parameters, the $Gumbel\ Softmax$ method in Eq.~(\ref{e11}) is used to find the sampling points in the training stage. In the inference stage, there is no need to consider whether the backpropagation is truncated. So the $argmax$ and $argsort$ functions can be directly used to find the sampling points and the points in local areas. Note that only the indices of sampling points are trained, because among all points in a local area, the features of a sampling point are the closest to the features of the local area. This can greatly reduce the storage and computation cost caused by differentiable indexing.

PSNet will update the abstract feature of the local areas to reflect the points that carry more meaningful information when the deep learning network performs the feature abstraction in this local area. On the contrary, the points with the weaker impact may be excluded from the local area.

\subsection{Supervision Information}
\label{sec:sm}

PSNet is embedded into a training network and co-trained with the training network, through which PSNet can be adapted to provide the data-structuring results more effective for the feature abstraction operators of the training network and the objectives of the learning task. PSNet does not have its own loss function. Instead, the training of PSNet is supervised by the loss function of the learning task running on the training network that PSNet is plugged into. 

For example, assuming that the loss function is cross-entropy, the loss function for a single sample (\ie a shape in a classification task or a point in a segmentation task) can be formulated as:

\begin{equation}
L_{task}=-\sum_{c=1}^{M}y_{c}log(p_{c})
\label{e12}
\end{equation}
where $M$ is the number of categories, $y_{c}$ is the label of the sample, $p_{c}$ is the prediction probability of the sample.

\subsection{Handling Large Scale Point Cloud Scenes}

Current mainstream deep learning models (\eg PointNet++, PointConv, RS-CNN, GAC, \etal) can only handle point clouds of limited scale (usually within 10K points). When being faced with larger-scale point cloud data (such as 1 million points), the input point clouds of the deep learning networks are typically pre-processed to reduce to a smaller scale. The following are the pre-processing methods used in the above mainstream deep learning models for processing shape classification tasks and the scene segmentation task.

Shape classification for a point cloud focuses more on the overall features of the object, the points are randomly sampled in the preprocessing phase to reduce the point cloud to an acceptable scale. The sampled points are then input into the deep learning networks. The time complexity of the random sampling pre-processing method is $O(1)$. 

As for the scene segmentation task, the unconstrained random sampling method may cause the loss of fine-grained features of a point cloud. Therefore, a large scale point cloud scene is first partitioned into regular voxels or grids. Then the random sampling is performed within each voxel or grid. This way, the fine-grained features of the point cloud can be preserved. 

\subsection{Time Complexity Analysis of PSNet}

The time complexities of FPS and kNN neighborhood query is $O(m^2)$ \cite{lang2020samplenet} and $O(ms+mlog_{2}n)$ (distance calculation and heap sort), respectively. The total time complexity of FPS+kNN is $O(m^2+ms+mlog_{2}n)$. The time complexity of a single-layer PSNet is $O(ms+mlog_{2}n)$ (\emph{SFTF} and heap sort). 

Considering that $s$ and $n$ are much smaller than $m$ in general, PSNet is expected to be much faster than FPS+kNN. Moreover, since \emph{SFTF} processes each point independently, it can be embarrassingly parallelized in modern deep learning frameworks on GPUs, which manifests the feature of weak scaling. Namely, the processing time for the point cloud remains constant as the number of points increases, and the speedup increases linearly as the number of processing elements increases. 

On the contrary, FPS cannot be parallelized because FPS has to query the points one by one in sequence. Therefore, at least $s$ iterations are required for subsampling. kNN calculates the Euclidean distance between each of the sampling points and all other points, which can be parallelized in theory. However, the unit operation in kNN is to calculate the Eulidean distance between two points while the unit operation in \emph{SFTF} is a multiplication, \ie a feature value times a model weight. The cost of unit operation in kNN is much higher than that of multiplication. Thus, even if kNN is fully parallelized, it is still much slower than our PSNet. We have conducted the experiments to verify this (Table \ref{t3}).

Furthermore, it is easy to implement PSNet. It does not require the developers to have advanced CUDA programming skills. It can be easily implemented on GPU using high-level programming paradigm provided in the deep learning frameworks such as TensorFlow \cite{abadi2016tensorflow} and PyTorch \cite{paszke2019pytorch}.

\section{Experiments}
\label{sec:Experiments}

We have conducted extensive experiments to evaluate PSNet. Our experiments consist of five parts: effectiveness, efficiency, robustness, ablation studies and visualization. In the experiments about effectiveness, we process the tasks in three mainstream application scenes of point clouds. We modified the deep learning models in the literature by replacing subsampling and  grouping in the original models with PSNet, and compared the performance between the original and the modified models. In the efficiency experiments, we recorded the inference time and the training time of the deep learning models with PSNet, and compared the time with the original models. In addition, we provided the visualization of the data structuring results.

Our comparison experiments strictly follow the experimental and hyperparameter settings of the original model. The basic model transformation such as translation, scaling, rotation and jitter are also randomly performed to enhance the robustness of model training.

\subsection{Effectiveness}

\subsubsection{Classification and Part Segmentation}

To evaluate the performance of PSNet in the application of shape classification and part segmentation, we modified the point-based method PointNet++ \cite{qi2017pointnet}, convolution-based methods PointCNN \cite{li2018pointcnn}, PointConv \cite{wu2019pointconv}, RS-CNN \cite{liu2019relation} and DensePoint \cite{liu2019densepoint} by replacing their data structuring methods with PSNet. The benchmark dataset of classification and segmentation is ModelNet40 \cite{wu20153d} and ShapeNet \cite{chang2015shapenet} respectively. The results are shown in Table~\ref{t1}. 

\begin{table}[t]
\begin{center}

\begin{tabular}{lcccccc}
\hline
\multirow{3}{*}{Methods} & \multicolumn{2}{c}{Classification}               & \multicolumn{4}{c}{Part Segmentation}                              \\
                         & \multirow{2}{*}{Orig.} & \multirow{2}{*}{PSNet} & \multicolumn{2}{c}{C-mIoU} & \multicolumn{2}{c}{I-mIoU} \\
                         &                           &                      & Orig.    & PSNet              & Orig.      & PSNet               \\ \hline
PN++(MSG)  & 91.9                      & \textbf{92.3}        & 81.9        & \textbf{82.1}    & 85.1          & 85.0              \\
PN++(SSG)  & 91.7                      & \textbf{92.2}        & 81.6        & \textbf{82.1}    & 84.8          & 84.8              \\
PointCNN         & 92.2                      & \textbf{92.3}        & 84.6        & \textbf{84.7}    & 86.1          & \textbf{86.2}     \\
PointConv        & 92.5                      & 92.4                 & 82.8        & \textbf{82.9}    & 85.7          & 85.7              \\
RS-CNN           & 92.6                      & 92.6                 & 84.1        & \textbf{84.2}    & 85.8          & 85.8              \\
DensePoint       & 93.2                      & \textbf{93.3}        & 84.2        & 84.2             & 86.4          & 86.4              \\ \hline
\end{tabular}

\end{center}
\caption{Effectiveness of deep learning models integrated with PSNet; The classification task is run on ModelNet40 and the shape part segmentation task run on ShapeNet, Orig. stands for Original Model, C-mIoU for class mIoU, I-mIoU for instance mIoU, PN++ for PointNet++, SSG for single scale grouping, MSG for multi-scale grouping.}
\label{t1}
\end{table}

In PointConv and RS-CNN, the performance of the network with PSNet is almost the same as that of the original model. There is the slight performance improvement compared to the original model in PointCNN and DensePoint with PSNet. Note that no matter it is single scale or multi-scale grouping, the accuracy of PointNet++ with PSNet is even better than that of the original models. This may be because PSNet is able to provide more appropriate local grouping than the original models. These results show that PSNet improves the effectiveness of the deep learning models.

\subsubsection{Scene Segmentation} The scene segmentation task is more complex than shape classification and part segmentation. It contains a large amount of noise data. The number of input points is very large, which can verify the effectiveness of PSNet in processing large-scale point clouds. We use the Stanford 3D Large-scale Indoor Spaces (S3DIS) \cite{floros2012joint} dataset in the experiments. The common scene segmentation benchmark is trained on areas 1-4 and tested on a completely independent area 5. We modified PointNet++ \cite{qi2017pointnet}, PointCNN \cite{li2018pointcnn} and the graph-based method GAC \cite{wang2019graph}, and conducted the comparison experiments. The experimental results are shown in the Table~\ref{t2}.

Compared with the original model, PSNet achieves better or the same effectiveness. The performance of PointNet++ with PSNet is even better than the original PointNet++ multi-scale grouping model. This result shows that PSNet is effective in building local graphs for the graph neural networks. GACNet strengthen the robustness of local changes by local relationships representation. This makes their effectiveness more difficult to benefit from stable data structure.

\begin{table}[t]
\begin{center}
\begin{tabular}{ccccc}
\hline
\multirow{2}{*}{Methods} & \multicolumn{2}{c}{mIoU}                                                                                             & \multicolumn{2}{c}{OA}                                                                                             \\
                         & original                                             & PSNet                                                           & original                                           & PSNet                                                           \\ \hline
PointNet++       & \begin{tabular}[c]{@{}c@{}}53.1\\ (MSG)\end{tabular} & \textbf{\begin{tabular}[c]{@{}c@{}}53.2\\ (SSG)\end{tabular}} & \begin{tabular}[c]{@{}c@{}}84.0\\ (MSG)\end{tabular} & \textbf{\begin{tabular}[c]{@{}c@{}}84.4\\ (SSG)\end{tabular}} \\
PointCNN         & 57.3                                                 & \textbf{57.4}                                                 & 85.9                                               & 85.9                                                          \\
GACNet          & 62.9                                                 & 62.9                                                          & 87.9                                               & 87.8                                                          \\ \hline
\end{tabular}
    
\end{center}
\caption{Scene segmentation comparison on S3DIS area5.}
\label{t2}
\end{table}

\subsubsection{Stability} 
\begin{figure}[t]
\begin{center}
\includegraphics[scale=0.8]{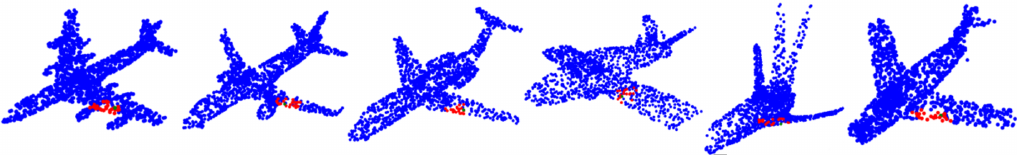}
\end{center}
   \caption{Visualization of stability of the data structuring results by PSNet in processing objects with similar features. In each shape, the red points represent a local area.}
\label{fig4}
\end{figure}

Stable data structuring methods help improve the effectiveness of the models. Compared with the widely used FPS+kNN method, PSNet is stable. We demonstrate this by randomly selecting some airplane shapes and processing them with PSNet. The results are visualized in \figurename~\ref{fig4}. 

A channel in PSNet represents an abstract spatial feature (the number of channels is set to be same as the number of local areas). The points with the highest scores in a channel are grouped as a local area (the point with the highest score in a channel is the sampled point), meaning that these points share the similar abstract spatial feature embedded in the channel. We randomly selected a local area no. (No. 16 in \figurename~\ref{fig4}). It can be seen from \figurename~\ref{fig4} that when processing different airplane models, local area No. 16 is always located at a similar part of the airplane (the part of airplane wing that is close to the airplane body). These results show that the grouping results produced by PSNet represent some particular geometric meaning (embedded in the corresponding channel). On the contrary, the local areas produced by FPS+kNN are random. If we run FPS multiple times even for the same point cloud model, the sampling results of different runs are different. If there are different point orderings for the same point cloud, the sampling results by FPS will be different too. The randomness of FPS and consequently the grouping results by kNN affect the consistency and effectiveness of the data structuring results. 

\subsection{Efficiency}
We conducted a series of experiments to verify the efficiency of PSNet. In this section, the MLP channels of the \emph{SFTF} $T$ in PSNet were set to 5, 32, 128 and $s$. All experiments were conducted on TITAN RTX. For fair comparison, all methods are implemented in PyTorch \cite{paszke2019pytorch}.

\subsubsection{Inference Time} We evaluated the efficiency of the forward propagation of PSNet, including the running time of PSNet and the overall running time of the entire training network which PSNet is plugged in. We compare the efficiency of PSNet with that of FPS and the networks using FPS. In the experiments, the number of points is 1024 or 2048 for the ordinary point clouds, and 30,000 or 80,000 for large scale point clouds. 

\begin{table}[t]
    \begin{center}

        \begin{tabular}{llllll}
        \hline
        \multirow{2}{*}{m→s} & \multicolumn{3}{c}{Inference Time(ms)} & \multicolumn{2}{c}{Memory}    \\
                             & FPS       & kNN    & PSNet              & F+k       & PSNet           \\ \hline
        1024→128             & 28.3      & 1.4     & \textbf{0.64}    & \textless{1M} & \textless{1M} \\
        1024→512             & 109.1     & 2.1     & \textbf{0.78}    & \textless{1M} & \textless{1M} \\
        2048→128             & 28.7      & 4.0     & \textbf{0.68}    & 1.5M          & 1.5M          \\
        2048→512             & 110.4     & 4.1     & \textbf{0.80}    & 8M            & \textbf{6M}   \\
        30000→512            & 128.0     & 30.4    & \textbf{0.81}    & 85M           & \textbf{75M}  \\
        30000→1024           & 235.4     & 30.7    & \textbf{0.80}    & 160M          & \textbf{140M} \\
        30000→4096           & 970.3     & 31.5    & \textbf{0.81}    & 633M          & \textbf{550M} \\
        80000→512            & 215.6     & 75.5    & \textbf{0.81}    & 270M          & \textbf{235M} \\
        80000→1024           & 420.1     & 77.1    & \textbf{0.81}    & 490M          & \textbf{440M} \\ 
        80000→4096           & 1577.3     & 80.4    & \textbf{0.90}    & 1960M          & \textbf{1735M} \\ 
        80000→16376           & 6701.1     & 97.3    & \textbf{0.97}    & 8340M          & \textbf{7144M} \\ \hline
        \end{tabular}

    \end{center}
    \caption{Inference time and memory in data structuring, $m$ and $s$ are the number of input points and sampled points respectively. Batch size is 8. F+k means FPS+kNN.}
    \label{t3}
\end{table}

\begin{table}[t]
\begin{center}
\setlength{\tabcolsep}{0.8mm}{
\begin{tabular}{llccc}
\hline
\multicolumn{1}{c}{Task}                                                                         & \multicolumn{1}{c}{Methods} & Original & S\&G(\%) & PSNet             \\ \hline
\multirow{5}{*}{\begin{tabular}[c]{@{}l@{}}Shape\\ Classification\\ (32/batch)\end{tabular}} & PointNet++          & 262.5    & 62.9     & \textbf{97.6}   \\
                                                                                                 & PointCNN            & 277.4    & 60.4     & \textbf{110.4}  \\
                                                                                                 & PointConv         & 335.4    & 61.2     & \textbf{131.2}  \\
                                                                                                 & RS-CNN             & 340.9    & 57.5     & \textbf{145.5}  \\
                                                                                                 & DensePoint         & 320.1    & 58.1     & \textbf{135.4}  \\ \hline
\multirow{5}{*}{\begin{tabular}[c]{@{}l@{}}Part\\ Segmentation\\ (32/batch)\end{tabular}}    & PointNet++          & 197.3    & 75.9     & \textbf{47.6}   \\
                                                                                                 & PointCNN            & 200.5    & 71.1     & \textbf{58.1}   \\
                                                                                                 & PointConv          & 210.9    & 72.6     & \textbf{58.1}   \\
                                                                                                 & RS-CNN             & 220.6    & 70.4     & \textbf{65.5}   \\
                                                                                                 & DensePoint         & 212.7    & 70.7     & \textbf{63.1}   \\ \hline
\multirow{3}{*}{\begin{tabular}[c]{@{}l@{}}Scene\\ Segmentation\\ (16/batch)\end{tabular}}   & PointNet++          & 468.7    & 46.1     & \textbf{256.7} \\
                                                                                                 & PointCNN            & 494.5    & 44.7     & \textbf{275.8}  \\
                                                                                                 & GACNet             & 510.1    & 44.1     & \textbf{279.4}  \\ \hline
\end{tabular}
}
\end{center}
\caption{Inference time (ms) spent in network forward propagation. S\&G(\%) is the percentage of time spent in sampling and grouping in the original models.}
\label{t4}
\end{table}

We recorded the time spent in sampling and grouping separately. But since in PSNet sampling and grouping are performed at the same time, we only recorded the total time spent by PSNet. We also recorded the memory consumption of the methods. The results are shown in the Table~\ref{t3}.

It can be observed from Table~\ref{t3} that with PSNet the inference time is reduced dramatically (becomes almost neglectable), while either the less or same amount of memory are consumed. FPS is sensitive to the number of sampling points. It is caused by the iterative implementation of FPS. The time of the kNN grouping increases only slightly as the number of points in the point clouds increases.

Moreover, the time spent by PSNet remains almost the same as the scale of the point clouds or the number of sampling points increases. This is because the calculations in PSNet can be embarrassingly parallelized and therefore can take full advantage of parallel capacity in GPU. Even in the smallest scales of the point cloud and sampling points, PSNet spends only 2\% of the time by FPS+kNN. 

In theory, the parallelized kNN should have the same execution time as the parallelized PSNet. However, it can be seen from Table~\ref{t3} that the time of PSNet is much less than that of kNN. Also, the time of PSNet remains almost constant as the numbers of input points and sampling points increase, while kNN does not. The reason for these may be because kNN mainly calculates the distances between points while PSNet processes the $1\times1$ convolution. The default parallelization schemes provided by the deep learning frameworks on GPU may be different for these two types of processing.

We also evaluated the inference time spent in network forward propagation. The results are shown in Table~\ref{t4}. The time spent in forward propagation of the network with PSNet is reduced significantly. For example, the time spent by the PointNet++ network with PSNet is only 24.1\% of the original time in the part segmentation task. Further, we notice that the reduction percentage is related to the complexity of the feature abstraction method in the network. If the method itself is more complex, FPS and local grouping will take a smaller proportion of time in forward propagation. This observation indicates that PSNet does not interfere with the feature abstraction process of the model.

\subsubsection{Training Time} As PSNet incorporates new training parameters in the model, it is a natural concern that it may lead to the increase in training time. We evaluate the training time, the results are shown in the Table~\ref{t5}.

\begin{table}[t]
\begin{center}
\small{
\begin{tabular}{llcc}
\hline
\multicolumn{1}{c}{Task}                                                                     & \multicolumn{1}{c}{Methods} & Original & PSNet          \\ \hline
\multirow{5}{*}{\begin{tabular}[c]{@{}l@{}}Shape\\ Classification\\ (32/batch)\end{tabular}} & PointNet++          & 104      & \textbf{68}  \\
                                                                                             & PointCNN            & 109      & \textbf{70}  \\
                                                                                             & PointConv          & 117      & \textbf{71}  \\
                                                                                             & RS-CNN             & 106      & \textbf{69}  \\
                                                                                             & DensePoint         & 101      & \textbf{66}  \\ \hline
\multirow{5}{*}{\begin{tabular}[c]{@{}l@{}}Part\\ Segmentation\\ (32/batch)\end{tabular}}    & PointNet++          & 188      & \textbf{118} \\
                                                                                             & PointCNN            & 193      & \textbf{125} \\
                                                                                             & PointConv          & 201      & \textbf{133} \\
                                                                                             & RS-CNN             & 194      & \textbf{122} \\
                                                                                             & DensePoint         & 184      & \textbf{114} \\ \hline
\multirow{3}{*}{\begin{tabular}[c]{@{}l@{}}Scene\\ Segmentation\\ (16/batch)\end{tabular}}   & PointNet++          & 1410     & \textbf{810} \\
                                                                                             & PointCNN            & 1433     & \textbf{811} \\
                                                                                             & GACNet             & 1445     & \textbf{815} \\ \hline
\end{tabular}
}
\end{center}
\caption{Comparison in network training time (ms).}
\label{t5}
\end{table}

Although PSNet introduces new parameters, the training time of the network with PSNet is less than that of the original network. The improvement in forward propagation is far greater than the extra time taken in deriving and updating the new parameters, which results in a substantial reduction in the overall training time.

\subsection{Comparison with Learning-based Sampling Methods}

In addition to comparing with the original models of the respective methods, we also compared PSNet with other learning-based sampling methods in literature: SampleNet\cite{lang2020samplenet} and CP-Net\cite{nezhadarya2020cpnet}. The data structuring part of PointNet++ is replaced by a learning-based method. However, SampleNet and CP-Net can only perform sampling. They have to be paired with other grouping methods such as kNN to complete data structuring. To the best of our knowledge, there is yet not a method which can perform sampling and grouping at the same time like PSNet. 

We used three metrics to evaluate the comparison: inference time for data structuring, accuracy for the classification task on ModelNet40, and class mIoU for the part segmentation task on ShapeNet. It can be seen from the Table~\ref{t55} that SampleNet and CP-Net take slightly longer time than PSNet even if they only perform sampling. When they are paired with the local gropuing method kNN, the total time spent by them in data structuring is 5.9 times that spent by PSNet. Although PSNet consumes less time, the performance in terms of both classification accuracy and segmentation mIoU is better than that of SampleNet and CP-Net. This is because PSNet can achieve adaptive sampling and grouping results through spatial features, which makes its data structuring results more suitable for both the feature abstraction and the specific objectives of the learning tasks. Although the sampling of SampleNet and CP-Net is adaptive to the tasks or the features to some extent, the paired heuristic grouping method (kNN) cannot improve the effectiveness of grouping.

\begin{table}[t]
    \begin{center}
    \setlength{\tabcolsep}{1.0mm}{
\begin{tabular}{lccccc}
\hline
Methods & Sampling   & Grouping   & Total & Cls. Acc. & Seg. mIoU \\ \hline
SampleNet & 1.04        & 4.10       & 5.14  & 91.8      & 81.6           \\
CP-Net    & 0.82        & 4.10       & 4.92  & 91.7      & 81.9           \\
PSNet       & \multicolumn{2}{c}{\textbf{0.80}} & \textbf{0.80}  & \textbf{92.2}      & \textbf{82.1}           \\ \hline
\end{tabular}
    }
    \end{center}
    \caption{Comparison with other learning-based methods; ``Sampling" and ``Grouping" are the inference time (ms) at the sampling and the grouping stage of data structuring respectively, ``Cls. Acc." and ``Seg. mIoU" are the accuracy of the shape classification task and class-mIoU of the part segmentation task respectively; $m$=2048, $s$=512, batch size=8.}
    \label{t55}
\end{table}

\subsection{Visualization}
\subsubsection{Visualization of subsampling}

\begin{figure}[ht]
\begin{center}
\includegraphics[width=3.4in]{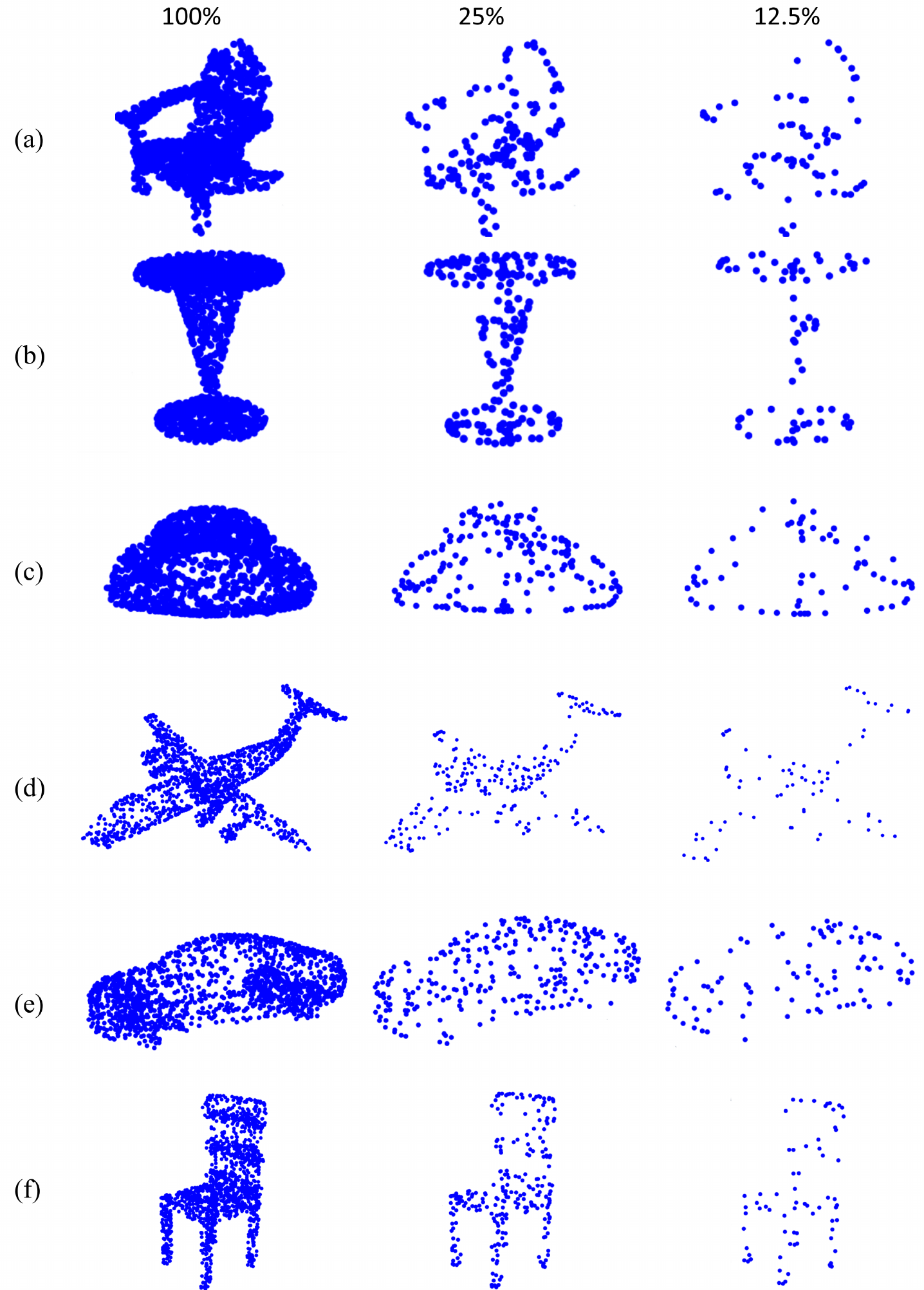}  
\end{center}
   \caption{Visualization of subsampling results; the subsampling rates are 25\% in the second column and 12.5\% in the third column.}
\label{fig5}
\end{figure}

In this subsection, we visualize the subsampling results during the PSNet process. The point whose features best match those of a local area is selected as a subsampling point (\ie the representative point of the local area). \figurename~\ref{fig5} shows the subsampling results on a few randomly selected point cloud shapes from ShapeNet when they are processed by PointNet++ plugged with our PSNet, where the point cloud in the first column is subsampled to 25\% and 12.5\% of the points in the second column and the third column, respectively. As can be seen from the figure, the points are uniformly sampled from the point cloud. Although the number of the points is greatly reduced after subsampling, the basic structures of the objects are still retained. These results demonstrate the excellent subsampling ability of PSNet.

\subsubsection{Visualization of local grouping on point cloud shapes}

\begin{figure}[ht]
\begin{center}
\includegraphics[width=3.4in]{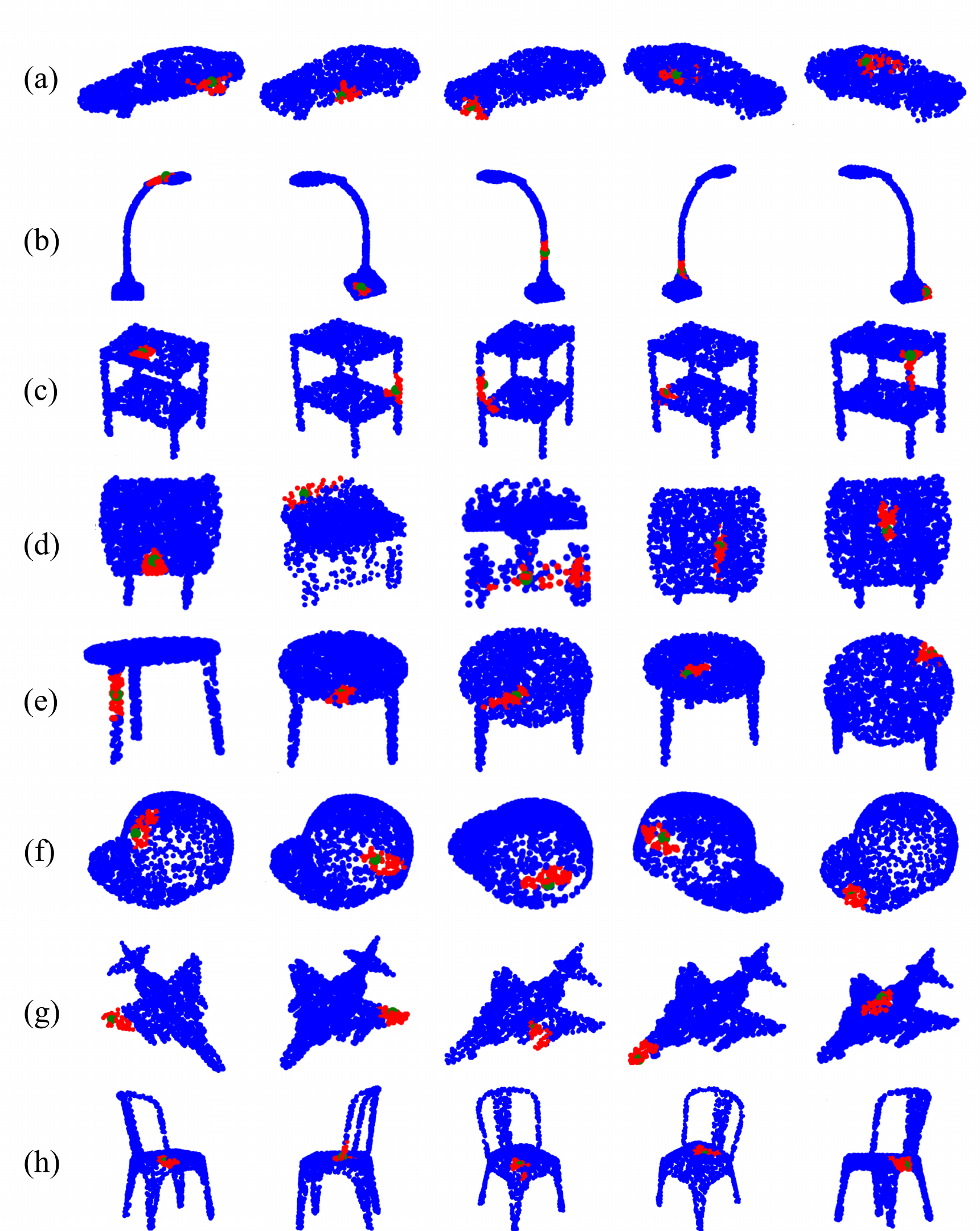} 
\end{center}
   \caption{Visualization of local grouping on point cloud shapes; a red area is a randomly selected local area after grouping.}
\label{fig6}
\end{figure}

In this section, we randomly select and visualize the local areas generated on the point cloud shapes from ShapeNet during the PSNet process. Their visualization results are shown in \figurename~\ref{fig6}. The red points are a local area grouped by PSNet. The green point in a local area is the point selected as the subsampling point. Note that the points are depicted according to their actual positions in the 3-D space and a green point may be partially covered by some red points. As can be seen from \figurename~\ref{fig6}, PSNet is able to effectively group the points with similar coordinates as a local area by learning from the input spatial features of individual points.

\begin{figure}[ht]
\begin{center}
\includegraphics[width=3.4in]{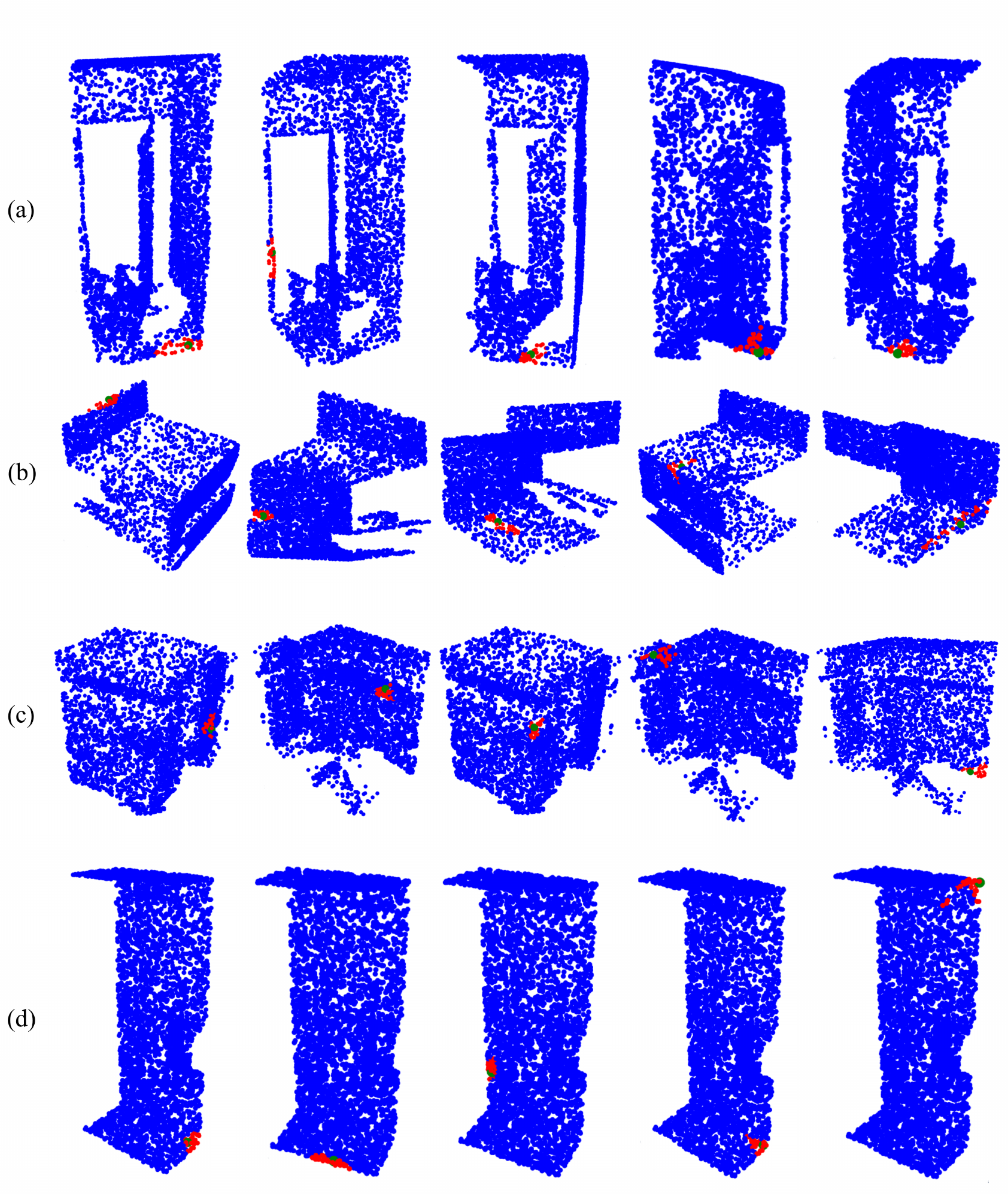} 
\end{center}
   \caption{Visualization of local grouping results for realistic data; a red area is a randomly selected local area after grouping.}
\label{fig7}
\end{figure}

\subsubsection{Visualization of local grouping on the point clouds generated from real scene}
The point cloud data generated by scanning the real scenes may contain a large amount of noise. Therefore it is more difficult to handle these realistic data in training. In this section, we randomly select and visualize the generated local areas of the scenes from the S3DIS dataset during the PSNet process. Their visualization results are shown in \figurename~\ref{fig7}. The coloring scheme in this figure is the same as that in \figurename \ref{fig6}. As can be seen from \figurename~\ref{fig7}, PSNet also works effectively for the realistic data.

\subsubsection{Visualization comparison with other methods}

In order to further understand the effectiveness of PSNet, in this subsection, we visualized the data structuring results obtained by PSNet and other sampling and grouping methods. 

\begin{figure}[ht]
\begin{center}
\includegraphics[width=3.4in]{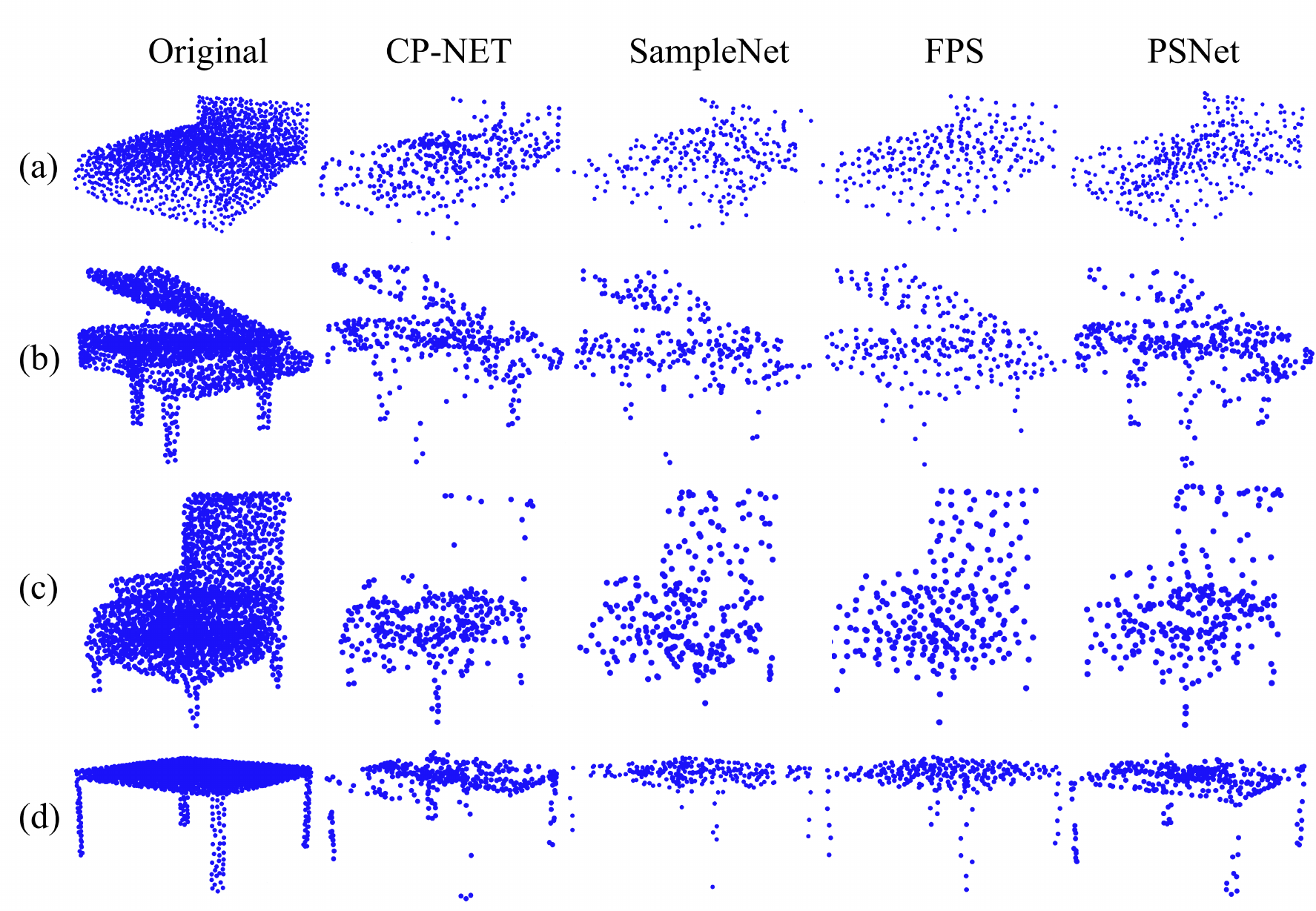} 
\end{center}
   \caption{The visualization comparison of the sampling results. The sampling rate is 25\%.}
\label{fig8}
\end{figure}

In particular, we visualized the sampling results obtained by the learning-based methods (\ie SampleNet and CP-Net) and the widely used uniform sampling method FPS. The experimental results are shown in \figurename~\ref{fig8}.

It can be seen that the points sampled by FPS are uniformly distributed, which causes FPS to equalize the points of different importance. For example, FPS sampled a large number of points for the bed base in \figurename~\ref{fig8}(a), the piano cover in (b), the chair back in (c) and the desktop in (d). However, these areas are simple and easy to describe. On the contrary, the contour of the bed headboard in (a), the legs of the piano in (b), the contour of the armrests and the legs of the chair in (c) and the legs of the table in (d) have fewer points.

The sampling result of SampleNet is closer to that of FPS, because its training process is based on the shape uniformly sampled by FPS. However, some features are missed in SampleNet, such as the edge of the bed headboard in (a), the left side of the keyboard in (b), the lower part of the chair legs in (c) \etal.

The sampling strategies of CP-Net and PSNet are similar since they both use the spatial features for sampling. Compared with FPS and SampleNet, CP-Net and PSNet reduce the number of the sampled points in the simple planes and retain the sampling points for the skeleton and the contour of the shapes. Compared with PSNet, CP-Net loses some shape details, such as the left edge of the chair back in (c), the middle of the table leg in (d) and so on.
In constrast, PSNet retains the points on the edge contour of the shapes and also the points that represent the skeleton and the detailed features of the shapes. This helps improve the effectiveness of the training model that PSNet is plugged into.

\begin{figure}[ht]
\begin{center}
\includegraphics[width=3.4in]{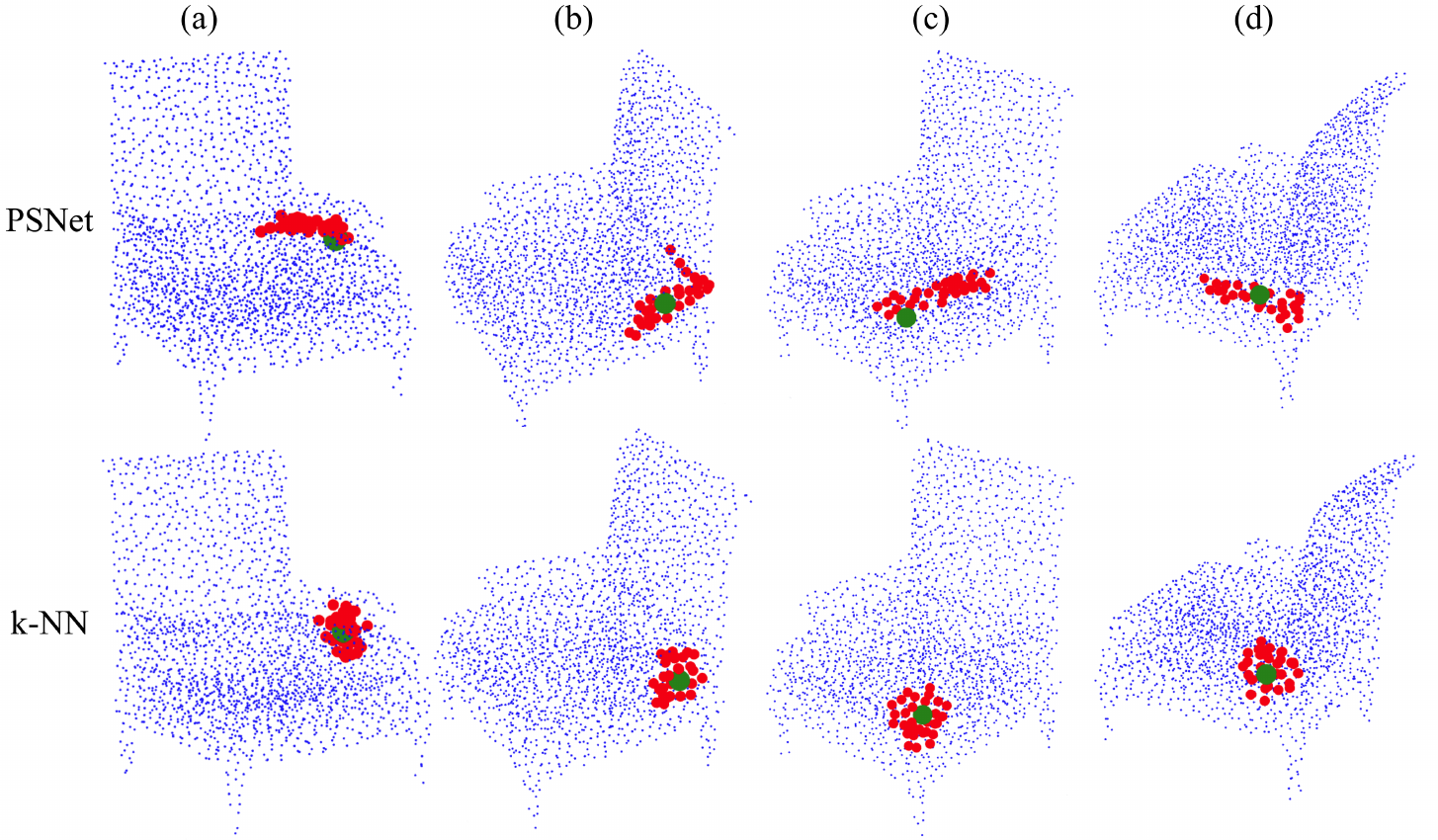} 
\end{center}
   \caption{The visualization comparison of the grouping results between PSNet and kNN. The green points are the sampling points of the local areas. The red points are the points in the local group.}
\label{fig9}
\end{figure}

In \figurename~\ref{fig9}, we visualized some grouping results achieved by PSNet and kNN. The grouping of kNN is based on the nearest points. The spatial feature correlation between the points in a group may be weak. Therefore, it is difficult for the grouped points to form a meaningful local shape. On the other hand, PSNet tends to allocate the points with similar spatial features into the same local area. For example, in \figurename~\ref{fig9}(a), the local grouping of PSNet is a local part of the chair cushion; in (b), the local grouping of PSNet is a right-angle shape of the seat cushion. However, the points in the seat cushion and the points in part of the armrest are allocated by kNN to the same local area; in (c) and (d), the local grouping of PSNet retains the linear contour feature of the edge of the seat cushion, while kNN mixes the points in the armrest, the seat cushion and the side panel.

\subsection{Robustness}

We evaluated the robustness of PSNet to point permutation and rigid transformation. We performed random permutation, translation and rotation on the test dataset and compared it with the original model. Table~\ref{t6} shows the results of running the shape classification task on RS-CNN with PSNet. The results show that the accuracy is not affected by the transformations. 

\begin{table}[hb]
\begin{center}
\setlength{\tabcolsep}{1.0mm}{
\begin{tabular}{lcccc}
\hline
                                                           & Original & Permutation & Translation & Rotation \\ \hline
\begin{tabular}[c]{@{}l@{}}RS-CNN \\ with PSNet\end{tabular} & 92.6     & 92.6        & 92.6        & 92.6     \\ \hline
\end{tabular}
}
\end{center}
\caption{Robustness experiment of PSNet.}
\label{t6}
\end{table}

\subsection{Ablation Experiments}

\subsubsection{The Effect of $\theta$ and $\varphi$}

\begin{figure}[t]
\begin{center}
\includegraphics[width=3.3in]{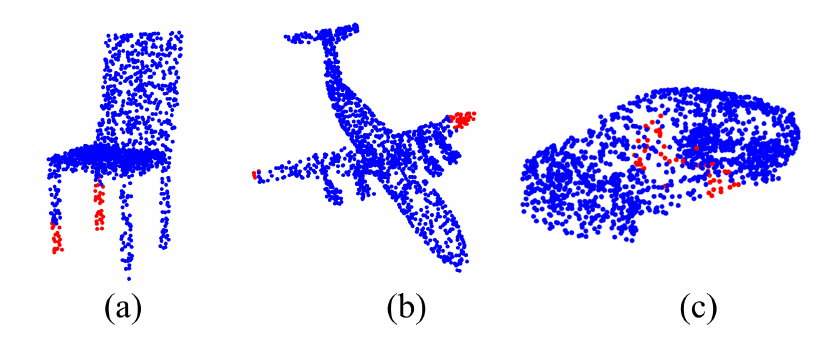} 
\end{center}
   \caption{Grouping error caused by the absence of $\theta$ and $\varphi$.}
\label{fig10}
\end{figure}

When only using the Cartesian coordinates, our PSNet may group some symmetric points, which are distant from each other, in a point cloud into the same local area. \figurename~\ref{fig10} visualizes this phenomenon. 

We evaluate the effectiveness of introducing $\theta$ and $\varphi$ in Eq.~(\ref{e6}) as the input spatial features of the points. Table~\ref{t7} shows the performance (in terms of class mean IoU and instance mean IoU) of PointNet++ with PSNet in the part segmentation task. The performance with other tasks and networks show the similar trend.

\begin{table}[hb]
\begin{center}
\begin{tabular}{lcc}
\hline
Spatial feature & \multicolumn{1}{l}{Class mIoU} & \multicolumn{1}{l}{Instance mIoU}\\ \hline
$\left[x,y,z\right]$             & 81.8       & 84.4          \\
$\left[x,y,z,\theta,\varphi\right]$           & \textbf{82.1}       & \textbf{84.8}          \\
$\left[r,\theta,\varphi\right]$               & 77.4       & 80.3          \\ \hline
\end{tabular}
\end{center}
\caption{The effectiveness of $\theta$ and $\varphi$ in PSNet. The part segmentation task was run on PointNet++ with PSNet.}
\label{t7}
\end{table}

\begin{table}[hb]
\begin{center}
\begin{tabular}{lccc}
\hline
 & \multicolumn{1}{l}{$x,y,z$} & \multicolumn{1}{l}{$x, y, z, \theta, \varphi$} & \multicolumn{1}{l}{$x, y, z, Q$}\\ \hline
error areas             & 174       & 4 &  5        \\
error points           & 1,431       & 5  & 7        \\
grouping error rate               & 4.47\%       & 0.02\%     &  0.02\%  \\ \hline
\end{tabular}
\end{center}
\caption{The grouping error rates with and without introducing $\theta$ and $\varphi$ in PSNet. 500 local areas are randomly selected from 20 symmetric shapes when running the part segmentation task on PointNet++ with PSNet. $Q$ is Quaternions. }
\label{t:error}
\end{table}

The results in Table~\ref{t7} show that even when only the Cartesian coordinates are used, PSNet is still very effective. But adding $\theta$ and $\varphi$ can further improve the effectiveness of PSNet. We examined the experimental records in detail. We found that when only using the Cartesian coordinates, PSNet may group some symmetric points with long distances into the same local area. We randomly checked 500 local areas of 20 symmetric shapes, and recorded the number of local areas which contain the points from distant symmetric parts (called \textit{error areas}) and the number of points which are incorrectly grouped (called \textit{error points}), and then calculated the grouping error rate. The values of the above records are listed in Table \ref{t:error}. As can be seen from the table, when only the Cartesian coordinates are used as the input spatial features (\ie the ``$x,y,z$" column in Table \ref{t:error}), there are 174 error areas (accounting for 34.8\% of all 500 local areas), containing 1,431 error points. There are 64 points in each local area and 500*64=32000 points in all 500 areas. So the grouping error rate is 1431/32000=4.47\%. After introducing $\theta$ and $\varphi$ as two extra input spatial features, the errors are almost eradicated as shown in the table. In the experiments with $\theta$ and $\varphi$ as additional input spatial features, however, the proportion is less than 1\%. We also conducted the experiments with only spherical coordinates as the input spatial features. The results show that the effectiveness is significantly reduced (last row of Table~\ref{t7}). The reason may be because it is easier to learn the features from the Cartesian coordinates. In addition, the $x, y, z, Q$ column in Table~\ref{t7} presents the error rate data for Quaternions \footnote{https://en.wikipedia.org/wiki/Quaternion} as an additional feature. The experimental results show that the effectiveness of Quaternions is similar to spherical coordinate. We used the spherical coordinate in the end because we think it is more intuitive.

\subsubsection{Effectiveness of Sampling and Grouping} 

In this section, we conducted the experiments to evaluate the effectiveness of sampling and grouping in PSNet. We integrated PSNet into PointNet++ by replacing the original FPS sampling and ball query grouping in PointNet++ with our PSNet. We used PSNet-integrated PointNet++ to process shape classification and part segmentation tasks. PSNet performs sampling and local grouping at the same time. In order to evaluate the effectiveness of PSNet in sampling and local grouping separately, we also conducted the experiments in which PointNet++ only used PSNet for sampling, but still used its original ball query for local grouping. The experimental results are shown in Table~\ref{t9}.

\begin{table}[t]
\begin{center}
\begin{tabular}{llcc}
\hline
\multirow{2}{*}{Data Structuring} & Classification                 & \multicolumn{2}{c}{Part Segmentation}          \\
                                  & Accuracy(\%)                   & Class mIoU                     & Instance mIoU \\ \hline
\begin{tabular}[c]{@{}l@{}}FPS + ball query\end{tabular}          & 91.7                           & 81.6                           & 84.8          \\
\begin{tabular}[c]{@{}l@{}}PSNet(Sampling)\\ +ball query\end{tabular}          & 91.9                           & 81.8                           & 84.8          \\
PSNet                               & \textbf{92.2} & \textbf{82.1} & 84.8          \\ \hline
\end{tabular}
\end{center}
\caption{The effectiveness of Grouping in PSNet; the PSNet is embedded into PointNet++; the data set for the shape classification task is ModelNet40, and the data set for part segmentation is ShapeNet.}
\label{t9}
\end{table}

It can be seen from the first two rows of Table~\ref{t9} that compared with the traditional sampling and grouping method (\ie FPS+ball query), the combination of PSNet sampling and ball query achieves better performance. This indicates that the adaptive sampling of PSNet is more effective than FPS. By comparing the second and the third row of Table~\ref{t9}, we can see that PSNet achieves better performance than the combination of PSNet sampling and ball query, which suggests that PSNet grouping is more effective than ball query. This result supports our argument that using the Euclidean distance (such as in ball query) as the only metric for grouping decisions may not be the best solution. Our PSNet can adjust the division of local areas adaptively based on local features, rather than on a single heuristic metric such as the distance, which we believe is the reason why PSNet achieved better performance. 

\subsubsection{The Channels of MLP} 
We compare the impact of the number of channels and the number of layers in MLP on the effectiveness of PSNet. Table~\ref{t8} shows the performance of running the part segmentation task with PointNet++ with PSNet. The performance with other tasks and networks show the similar trend. The results show that PSNet implemented with only one layer MLP (\ie $\left[5,32,s\right]$) is already effective. Increasing to two layers ($[5, 32, 128, s]$) can further improve the effectiveness slightly. But increasing the number of channels or layers further does not bring further benefit.

\begin{table}[t]
\begin{center}
\begin{tabular}{lcc}
\hline
Channels of MLPs & Class mIoU    & Instance mIoU \\ \hline
$\left[5,32,s\right]$                & 82.0            & 84.6          \\
$\left[5,32,128,s\right]$                 & \textbf{82.1} & \textbf{84.8} \\
$\left[5,64,256,s\right]$                 & \textbf{82.1} & \textbf{84.8} \\
$\left[5,32,128,256,s\right]$                  & \textbf{82.1} & \textbf{84.8} \\ \hline
\end{tabular}
\end{center}
\caption{The impact of the number of channels and the number of layers in MLP; the part segmentation task is run on PointNet++ with PSNet.}
\label{t8}
\end{table}

\section{Conclusion}
\label{sec:Conclusion}

In this paper, we propose a novel and fast data structuring method called PSNet for deep learning of point clouds. The way in which PSNet structures the point cloud data can significantly improve the training and inference speed without affecting the accuracy of the original model. The sampling and grouping in PSNet can be embarrassingly parallelized. Moreover, the sampling and grouping are performed at the same time in PSNet, while the current mainstream methods perform sampling and grouping in sequence as two separate processes. Because of these features, PSNet can work for the deep-learning of larger scale point clouds than those in literature. Moreover, PSNet can work with the deep learning networks in a plug and play manner. We believe PSNet can also be applied to various non-Euclidean data structures other than point clouds, which we plan to investigate in future.

\appendices
\section{Using PSNet in a Plug and Play Fashion}
\label{sec:pnp}

\begin{figure*}[hb]
\begin{center}
\includegraphics[width=6.7in]{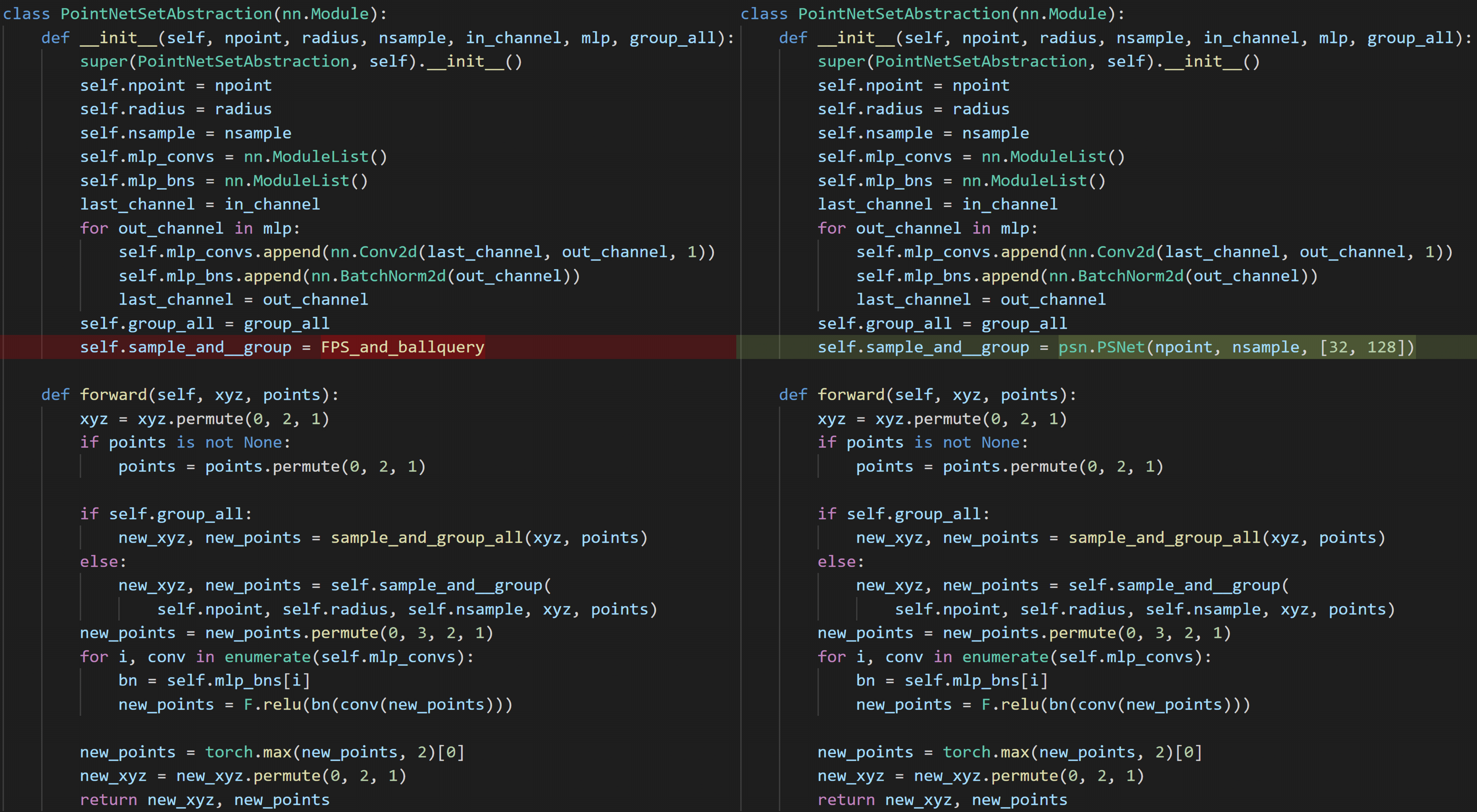}  
\caption{An example of using the plug and play PSNet in PointNet++.}
\label{pnp}
\end{center}
\end{figure*}

In this section, we present how we can plug and play PSNet in an existing point cloud deep learning model, PointNet++. The reason why we use the modification of PointNet++ as the example is because most hierarchical point cloud deep learning models are based on the PointNet++ architecture. 

\figurename~\ref{pnp} is the screen capture of the code snippet that shows how to replace the original data structuring method in the PyTorch implementation of PointNet++ with our PSNet. Only a line of code (the highlighted line) has to be modified to plug PSNet.

In the code snippet in \figurename~\ref{pnp}, Class $PointNetSetAbstraction$ is the main part of the feature abstraction component of the model. Each feature abstraction layer will construct an instance of this class. $PointNetSetAbstraction$ is mainly composed of three parts: data structuring, feature transformation and feature aggregation. In the original model, the data structuring method (\ie $sample\_and\_group$ in the highlighted line) is FPS (for subsampling) and ball query (for local grouping). The module lists, $mlp\_convs$ and $mlp\_bns$, are the operator sets for feature transformation. The \textit{max} function is used for feature aggregation. When plugging PSNet, the data structuring method in our PSNet module (\ie $psn.PSNet()$) is assigned to the $sample\_and\_group$ method. The input parameters in $psn.PSNet()$ are used to initialize the PSNet network. The parameter $npoint$ specifies the number of sampling points, which is also the number of local areas (\ie $s$ in the main paper), $nsample$ specifies the number of points in a local area (\ie $n$ in the main paper), the parameter $[32, 128]$ means that there are two hidden layers in the PSNet network, and the numbers of channels in these two layers are 32 and 128, respectively. Note that the numbers of channels in the input and output layers of the PSNet network are 5 (5 input spatial features) and $s$, respectively.  

Note that when PSNet is assigned to $sample\_and\_group$, the interface of invoking $sample\_and\_group$ remains the same although the input parameter $radius$ (\ie the radius of the ball query) in $sample\_and\_group$ is actually not needed (which is the parameter for the ball query). This is on purpose so that the programmers only need to change one line of code in the original training model for using PSNet. 

\section*{Acknowledgment} 
\addcontentsline{toc}{section}{Acknowledgment}

This work was support in part by the National Key R\&D Program of China under Grant 2018YFB2101504, in part by the National Natural Science Foundation of China under Grant 61672473, in part by the Key Research and Development Program of Shanxi Province of China under Grant 201803D121081 and Grant 201903D121147, and in part by the Natural Science Foundation of Shanxi Province of China under Grant 201901D111150.


\bibliographystyle{IEEEtran}
\bibliography{IEEEabrv,PSNet}

\end{document}